\documentclass[runningheads]{llncs}
\usepackage[T1]{fontenc}
\usepackage{multirow}
\usepackage{booktabs}
\usepackage{amsmath}
\usepackage{amsfonts}
\usepackage{array}
\usepackage{subcaption}
\usepackage{xurl}

\usepackage{graphicx, verbatim}
\begin{document}
\title{Invisible Attributes, Visible Biases: Exploring Demographic Shortcuts in MRI-based Alzheimer's Disease Classification}

%
\titlerunning{Invisible Attributes, Visible Biases}

\author{Akshit Achara \and
Esther Puyol Anton \and
Alexander Hammers \and 
Andrew P. King,  for the Alzheimer’s Disease Neuroimaging Initiative}

\authorrunning{Achara et al.}
%
\institute{School of Biomedical Engineering and Imaging Sciences, King's College London, UK}


%

\maketitle              
\begin{abstract}
Magnetic resonance imaging (MRI) is the gold standard for brain imaging. Deep learning (DL) algorithms have been proposed to aid in the diagnosis of diseases such as Alzheimer's disease (AD) from MRI scans.
However, DL algorithms can suffer from shortcut learning, in which spurious features, not directly related to the output label, are used for prediction. When these features are related to protected attributes, they can lead to performance bias against underrepresented protected groups, such as those defined by race and sex.
In this work, we explore the potential for shortcut learning and demographic bias in DL based AD diagnosis from MRI. We first investigate if DL algorithms can identify race or sex from 3D brain MRI scans to establish the presence or otherwise of race and sex based distributional shifts. Next, we investigate whether training set imbalance by race or sex can cause a drop in model performance, indicating shortcut learning and bias. Finally, we conduct a quantitative and qualitative analysis of feature attributions in different brain regions for both the protected attribute and AD classification tasks. Through these experiments, and using multiple datasets and DL models (ResNet and SwinTransformer), we demonstrate the existence of both race and sex based shortcut learning and bias in DL based AD classification. Our work lays the foundation for fairer DL diagnostic tools in brain MRI. The code is provided at \url{https://github.com/acharaakshit/ShortMR}

\end{abstract}
\section{Introduction}
Deep learning (DL) algorithms can suffer from shortcut learning, in which spurious correlations between input features and output labels are learnt, regardless of the features' actual relevance to the task~\cite{mehrabi2021survey,geirhos2020shortcut}.
When the features are associated with protected attributes such as race and sex these shortcuts can lead to bias. The biases can be further exacerbated by imbalanced representation of protected groups in datasets used for training DL models~\cite{barocas2023fairness,buolamwini2018gender}. Datasets such as Waterbirds \cite{sagawa2019distributionally} and CelebA~\cite{liu2015deep} for computer vision, and CheXpert~\cite{irvin2019chexpert} in medical imaging have been created for investigating such spurious correlations.

Spurious features can be learned by DL algorithms for tasks such as classification~\cite{oakden2020hidden,winkler2019association}, segmentation~\cite{moayeri2022hard} or generation~\cite{friedrich2024auditing,girrbach2025large}. For datasets such as Waterbirds~\cite{sagawa2019distributionally}, it is visually possible to identify the spurious correlations between the background (land and sea) and objects (land and sea birds). However, in many medical images, it is difficult or impossible for humans to identify the associated protected attribute(s), such as race and sex, i.e. it could be said that they are ``invisible'' to humans. However, DL algorithms can often detect them~\cite{gichoya2022ai,stanley2022fairness}. If a DL algorithm can distinguish between different protected groups, then there is the potential for features associated with these attributes to act as shortcuts, or to introduce ``visible'' biases in clinical classification tasks. 
Therefore, in this paper, we consider the question: ``can protected attribute based shortcut learning lead to bias in DL classification models for brain MRI scans?''

We first study if demographic attributes, namely race and sex, can be identified from structural 3D brain MRI scans on three different datasets. Second, we create baseline and ``biased'' datasets with differing levels of race/sex imbalance to study shortcut learning in AD classification from brain MRI. Finally, we perform a quantitative and qualitative analysis of regional feature attributions 
to reveal the nature of the shortcut learning and 
lay the foundations for future research on understanding the effect of these ``invisible'' protected attributes on ``visible'' biases.

\subsubsection*{Related Work}
In~\cite{stanley2022fairness}, the authors showed that sex can be classified from structural adolescent brain MRIs. However, the focus of this work was on assessing sex classification fairness rather than fairness in diagnostic tasks. Previous studies have shown the effect of protected group representation (sex) for AD classification on performance~\cite{petersen2022feature,bercea2023bias,wang2023bias,klingenberg2023higher}. However, the datasets were not specifically curated for assessing shortcut learning. In~\cite{stanley2025and}, the authors studied the effects of various synthetic biases on model performance using a similar strategy of dataset curation. However, the approach utilised synthetic data and synthetically introduced biases with a different research objective. 

\subsubsection*{Contributions}
We show that race and sex can be identified from 3D brain MRI scans on three different datasets  using two different DL models (ResNet~\cite{he2016deep} and SwinTransformer~\cite{liu2021swin}).
We construct baseline and biased datasets based on sex and race to highlight shortcut learning in DL based AD classification. Finally, we quantitatively and qualitatively analyse feature attributions to show that both sex and race can cause shortcut learning and bias in AD classification, as well as provide insight into the nature of the shortcuts.


\section{Methods and Experiments}
\label{experiments}

\subsubsection*{Notation}
\label{notation}
We use the following notation throughout this paper:

\begin{enumerate}
    \item[$\bullet$] $X \in \mathbb{R}^{N \times 1 \times D \times H \times W}$: MRI input images ($N$: number of samples; single channel; $D, H, W$: depth, height, width). $X_i\in\mathbb{R}^{1 \times D \times H \times W}$ represents a single input image, $i\in \{1,..,N\}$.
    \item[$\bullet$] $A \in \{0,1\}^N$: Binary protected attributes (e.g., sex, race) with $A_0, A_1$ denoting distinct protected groups (e.g., male/female, Black/White).
    \item[$\bullet$] $Y \in \{0,1\}^N$: Diagnostic labels with $Y_0, Y_1$ denoting specific diagnostic classes (i.e. CN: Cognitively normal, AD: Alzheimer's disease).
    \item[$\bullet$] $f$: Predictive classifier; either $f: X \rightarrow A$ (protected attribute classification) or $f: X \rightarrow Y$ (diagnosis).
    \item[$\bullet$] $T: X\to X'$ indicates a non-linear transform in image space.
    \item[$\bullet$] $L\in \mathbb{R}^{N \times 1 \times D \times H \times W}$ indicates the positive GradCAM attributions for $X$.
    \item[$\bullet$] $\Omega\in \{1,..,\omega\}^{1 \times D \times H \times W}$ represents an atlas map with $\omega$ regions.
    \item[$\bullet$] $R: \mathbb{R}^\omega\to\{1,...,\omega\}$ represents the attribution rank of each region in the atlas.
\end{enumerate}

\subsubsection*{Datasets}
\label{datasets}
\begin{enumerate}
    \item \textbf{ADNI}: The Alzheimer's Disease Neuroimaging Institute (ADNI)~\cite{petersen2010alzheimer} has released several data studies called \textit{ADNI 1}, \textit{ADNI GO} and \textit{ADNI 2} which consist of structural MRIs, acquisition parameters and protected attributes associated with the subjects. We perform skull stripping using HD-BET~\cite{isensee2019automated} on the already-preprocessed images for all our experiments. Each subject may have one or more images. This dataset was used in Experiment 1 on protected attribute classification and Experiment 2 on shortcut learning.
    \item \textbf{OASIS-3}: The Open Access Series of Imaging Studies 3 (OASIS-3)~\cite{lamontagne2019oasis} dataset consists of structural MRIs and protected attribute information for a combination of healthy subjects and subjects at different stages of cognitive decline. We select the bias-normalised images produced by Freesurfer~\cite{fischl2012freesurfer} and perform skull stripping on these images. Each subject may have one or more images. This dataset was used only in Experiment 1 on protected attribute classification.
    \item \textbf{HCP}: The Human Connectome Project (HCP)~\cite{van2013wu} consists of $1114$ structural MRIs from healthy subjects along with protected attribute information. We use the already preprocessed and skull-stripped images available with the dataset. Each subject has one sample in this dataset. This dataset was used only in Experiment 1 on protected attribute classification.
\end{enumerate}
All images were retained in native space and resized to $256\times256\times256$. Inputs were z-score normalised. Analyses were limited to adult MRIs from Black and White subjects due to limited data from other racial groups.

\subsubsection*{Experiment 1: Protected Attribute Classification}
\label{protectedattributeclassification}
The first experiment aimed to train and evaluate DL models for race and sex classification. Each classification task is binary, i.e. male/female and Black/White.
Formally, the protected attribute classification task can be defined as: $f: X\to A$. We use a stratified train-test ratio of $80:20$ for both race and sex classification tasks with validation sets being $8\%$ and $12\%$ of the training sets respectively. A smaller validation set is used for race classification due to the class imbalance. Stratification is based on race, sex, and age, with subjects categorized as `younger' or `older' using a single age threshold based on absolute age range for each dataset. One model was trained on each dataset.

\subsubsection*{Experiment 2: Shortcut Learning}
\label{protectedattributesasshortcuts}
In Experiment 2, we consider the CN vs. AD classification task, which can be formally defined as: $f:  X\to Y$, where $Y\in \{0,1\}^N$ are the diagnostic labels (i.e. CN and AD). We created two types of dataset: ``baseline'' and ``biased'', with chosen levels of (im)balance between protected attributes ($A_0, A_1$) and diagnostic labels $(Y_0, Y_1)$. In a baseline dataset, the combinations of $A\in{A_0,A_1},\ Y\in{Y_0, Y_1}$ are proportionally represented for all the samples $(X,Y)$, preserving their overall distribution within each diagnostic class.
In a biased dataset, the training set consists of majority samples $(X,Y)$ such that $A=A_1,\ Y=Y_0$ or $A=A_0,\ Y=Y_1$ and minority samples $(X,Y)$ such that $A=A_0,\ Y=Y_1$ or $A=A_1,\ Y=Y_0$. Here, $A_0$ and $A_1$ represent the two protected groups and $Y_0$ and $Y_1$ could be any of the two classes (CN or AD). The test set has an opposite notion of majority and minority samples, i.e. the majority and minority samples in the training set and the minority and majority samples in the test set. This imbalance in samples based on protected groups introduces a group-based distributional shift between training and test datasets.

\begin{figure}
\centering
  \includegraphics[width=\linewidth]{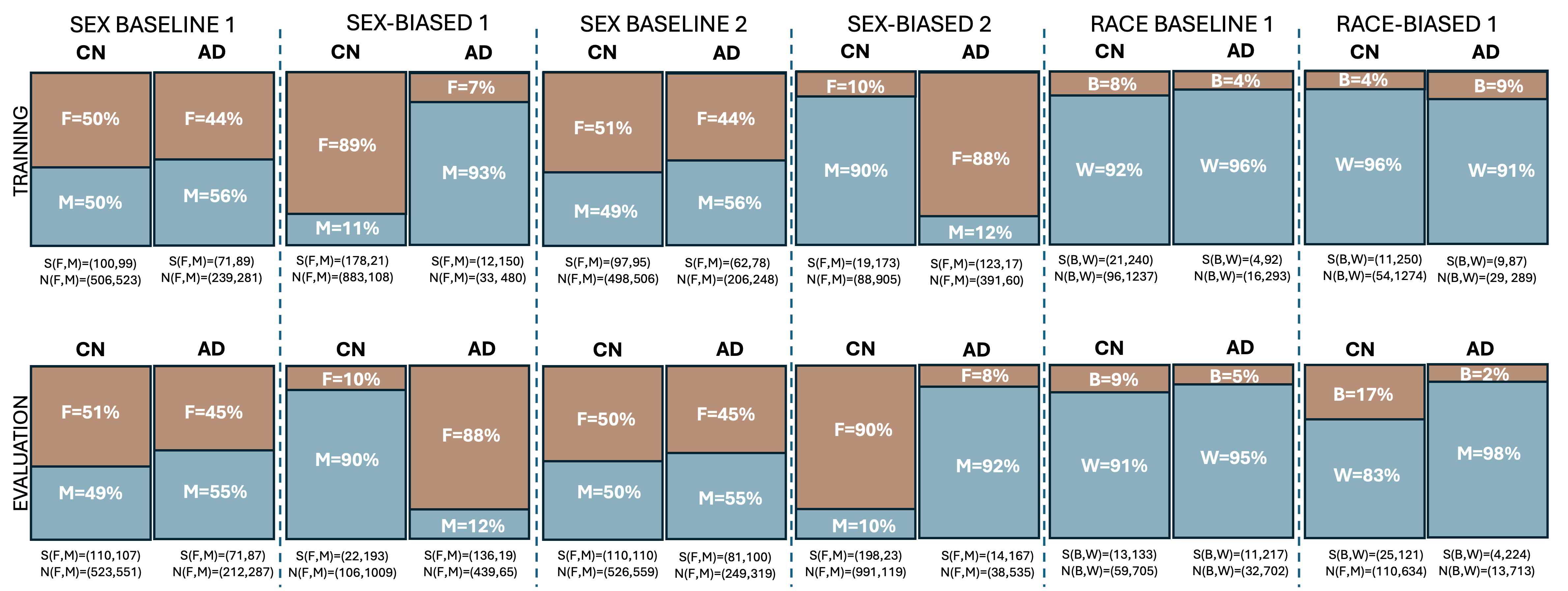}
  \caption{The formation of baseline and biased datasets. For the baseline datasets, the sex or race (im)balance between training and test sets are approximately the same. In the biased datasets they are different. Here, $S(A_0, A_1)$ indicates the number of subjects for each protected group $A_0$ and $A_1$. Similarly, $N(A_0, A_1)$ indicates the number of samples (i.e. images) for each protected group. $M,\ F,\ B$ and $W$ correspond to male, female, Black and White.}
  \label{fig:standard}
\end{figure}

The proportions of samples in the baseline and biased training and test datasets for each experiment can be seen in Figure~\ref{fig:standard}. The proportions of AD and CN subjects between the baseline and biased datasets are constrained to be equal (with a minor floating point error) to ensure fair comparison of models. For both balanced and biased datasets, we use a stratified validation set which is comprised of $10\%$ of the training data. The stratification is performed based on cognitive state, race, sex and age (`younger' and `older').
It is to be noted that there is a less severe difference in protected attribute imbalance between the race baseline and biased datasets due to the insufficient number of Black subjects. Additionally, we have not created a race biased dataset 2 due to lack of sufficient training data. 

\subsubsection*{Experiment 3: Interpretability}
\label{explainingpredictions}
We produce feature attribution maps to visualise the regions of the MRI scans that are used by the models in making their classifications. The interpretability method we use is  GradCAM~\cite{selvaraju2017grad} which is a layer based attribution method.  We use stability certification~\cite{jin2025probabilistic} to analyse the stability of feature attributions. This computes the probability that the prediction remains unchanged when the input with only patches consisting of top features (obtained from any interpretability method) is perturbed by up to $r$ features where $r$ is referred to as the radius. 
To enable quantification of attributions for different brain regions, the $L$ are transformed to $L'$ using $T$, which was computed by registering the $X_i$'s to the Hammersmith atlas~\cite{hammers2003three} via image registration. This is followed by obtaining the mean attribution rank vector over all $N$ test samples, $\textbf{r} = \dfrac{1}{N}\sum_{i=1}^{N} R(\{\dfrac{\sum L_i'\cap \Omega_1}{|\Omega_1|},...,\dfrac{\sum L_i'\cap \Omega_\omega}{|\Omega_\omega|}\})$. Here, $L_i'$ represents the GradCAM attribution map for sample $i$ in the atlas space, $\Omega_j$ represents all the pixels in region $j$ of the atlas and $L_i'\cap \Omega_j$ represents the pixels in $L_i'$ that fall in $\Omega_j$.
Effectively, we rank the mean attributions of the atlas regions for each sample, and then average the rank vectors over all samples.

Feature attribution maps and ranking vectors are produced in this way for both protected attribute classification models  ($\textbf{r}^{PA}$) and the biased and baseline diagnosis models ($\textbf{r}^{BI}$, $\textbf{r}^{BA}$). We then compute difference vectors $\textbf{B} = \textbf{r}^{BI} - \textbf{r}^{BA}$ and $\textbf{P} = \textbf{r}^{PA} - \textbf{r}^{BA}$ to find the regions most associated with bias and protected attributes, respectively. To quantify the extent of the shortcut learning we compute the Spearman's rank correlation coefficient between these two vectors. To reveal the regions most associated with shortcut learning we visualise the most significant shared regions netween $\textbf{B}$ and $\textbf{P}$ in the atlas space. We use one sample per subject for these interpretability experiments.
Overall, this approach is used to understand the relation between features from different models in the atlas space. These rank correlations can be interpretable metrics that provide insights into the occurrence of shortcut learning as well as quantifying its impact.

\subsubsection*{Implementation Details}
\label{implementationdetails}
We use two DL models in our experiments, namely SwinTransformer (with Convolutional Layers from MONAI~\cite{cardoso2022monai} and a window size of $5$) and ResNet50. Pretrained weights~\cite{deng2009imagenet} for ResNet50 were utilised for faster convergence due to the relatively small training datasets. Validation performance was used for early stopping mechanisms and obtaining best checkpoints based on validation set loss and F1-scores. It was ensured for all tasks that there was no subject overlap between the training, validation and test subsets. We used inverse probability weighting along with Cross Entropy\cite{shannon1948mathematical} as our loss function and a cosine learning rate scheduler with the AdamW\cite{loshchilov2017decoupled} optimiser. We used a batch size of $4$ for all tasks along with gradient accumulation. The code for the implementation is provided at \url{https://github.com/acharaakshit/ShortMR}.

\section{Results}
\label{results}
\subsubsection*{Experiment 1: Protected Attribute Classification}
\label{protectedattributeclassificationresults}
 Table~\ref{Tab:sexclassification} shows that both SwinTransformer and ResNet50 models have high F1-scores in classifying both male and female subjects. Similarly, in Table~\ref{Tab:raceclassification}, even with severe data imbalance, it can be seen that both models can classify White and Black subjects with a high level of accuracy.
\begin{table}
\centering
\setlength\abovecaptionskip{2pt}
\setlength\belowcaptionskip{3pt}
\vspace*{-5pt}  
\caption{Experiment 1 - Protected attribute classification. Performance measured by F1 score for SwinTransformer and ResNet50 models, for both race and sex classification across three datasets. The numbers in the cell (a,b) indicate the number of subjects in the complete and test datasets repsectively.}
\label{Tab:attributeclassification}
\begin{subtable}[t]{0.48\textwidth}
\centering
\subcaption{Sex classification}
\label{Tab:sexclassification}
\resizebox{0.95\textwidth}{!}{%
\begin{tabular}{@{}c@{\hskip 1em}c@{\hskip 1em}c@{\hskip 1em}c@{\hskip 1em}c@{}}
  \toprule
  \textbf{Dataset} & \textbf{Class} & \textbf{SwinTransformer} & \textbf{ResNet50} \\
  \midrule
  \multirow{2}{*}{ADNI} 
  & Female $(723, 149)$ 
    & $0.87$ 
    & $0.91$ \\
  & Male $(922, 184)$
    & $0.89$ 
    & $0.93$ \\
  \cmidrule(lr){2-4}
  \multirow{2}{*}{OASIS-3} 
  & Female ($722, 145$)
    & $0.93$ 
    & $0.93$ \\
  & Male $(577, 115)$
    & $0.91$ 
    & $0.91$ \\
  \cmidrule(lr){2-4}
  \multirow{2}{*}{HCP} 
  & Female $(543, 108)$
    & $0.91$ 
    & $0.94$ \\
  & Male $(455, 92)$
    & $0.88$ 
    & $0.92$ \\
  \bottomrule
\end{tabular}}
\end{subtable}%
\hspace{\fill}
\begin{subtable}[t]{0.48\textwidth}
\centering
\subcaption{Race classification}
\label{Tab:raceclassification}
\resizebox{0.95\textwidth}{!}{%
\begin{tabular}{@{}c@{\hskip 1em}c@{\hskip 1em}c@{\hskip 1em}c@{\hskip 1em}c@{}}
  \toprule
  \textbf{Dataset} & \textbf{Class} & \textbf{SwinTransformer} & \textbf{ResNet50} \\
  \midrule
  \multirow{2}{*}{ADNI} 
  & White $(658, 132)$
    & $0.98$ 
    & $0.98$ \\
  & Black $(85, 17)$
    & $0.81$ 
    & $0.77$ \\
  \cmidrule(lr){2-4}
  \multirow{2}{*}{OASIS-3} 
  & White $(1101, 221)$
    & $0.98$ 
    & $0.99$ \\
  & Black $(198, 39)$
    & $0.84$ 
    & $0.91$ \\
  \cmidrule(lr){2-4}
  \multirow{2}{*}{HCP} 
  & White $(830, 167)$
    & $0.98$ 
    & $0.97$ \\
  & Black $(168, 33)$
    & $0.93$ 
    & $0.84$ \\
  \bottomrule
\end{tabular}}
\end{subtable}
\end{table}
\vspace*{-2pt}

\begin{figure}[t]
\centering
\setlength\abovecaptionskip{2pt}   
\setlength\belowcaptionskip{2pt}   
\vspace*{-4pt}
\resizebox{0.95\textwidth}{!}{%
\includegraphics[width=\linewidth]{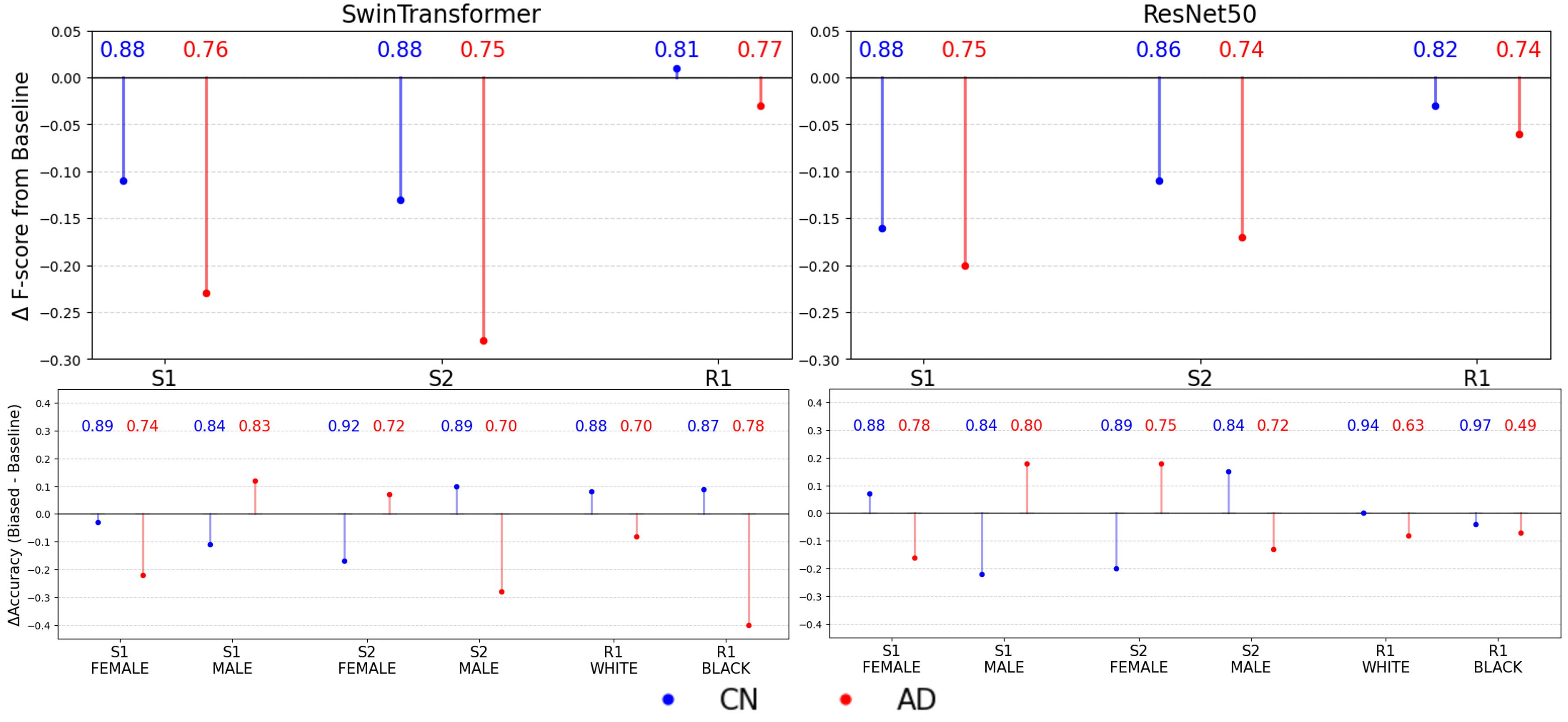}}
\caption{Experiment 2 - Shortcut learning. Figures show differences in performance (F1-score and accuracy) between models trained on baseline CN vs. AD task (from ADNI) and the same task with the corresponding biased datasets. $S1$, $S2$ and $R1$ represent the sex biased $1$, sex biased $2$ and race biased $1$ datasets respectively (see Figure \ref{fig:standard}). The number above each line indicates the baseline F1 score/accuracy, and a positive/negative $\Delta$ indicates an increase/drop in F1 score/accuracy from the baseline to the biased model. The first row shows overall class-level F1 scores and the second row shows protected group level accuracies.}
\label{fig:diseaseplots}
\end{figure}

\subsubsection*{Experiment 2: Shortcut Learning}
\label{shortcutresults}
Figure~\ref{fig:diseaseplots} shows the differences in class-level F1-scores between the baseline and biased models for the CN vs. AD task. A steeper drop is observed in the AD class as compared to CN. The second row highlights the differences in group-level accuracy scores between the baseline and biased models, depicting the occurrence of shortcut learning.

\subsubsection*{Experiment 3: Interpretability}
\label{explanationsresults}


For ResNet50, the Spearman's rank correlations $\rho$ between R(\textbf{B}) and R(\textbf{P}) (see Section~\ref{experiments}) were $0.85, 0.59$ and $0.32$ for the Sex 1, Sex 2 and Race 1 experiments, respectively (see Figure \ref{fig:standard}). Similarly, $\rho$ was $0.66$, $0.67$ and $0.08$ for the Sex 1, Sex 2 and Race 1 experiments, respectively for SwinTransformer. All $p$ values were $< 0.001$, indicating significant correlations except for the Race 1 experiment for SwinTransformer ($p=0.41$), where we did not see a correlation. We also performed a permutation test to account for dependence of \textbf{P} and \textbf{B} on \textbf{r}$^{BA}$. We observed significant p-values ($<0.05)$ for $S1$ using ResNet50 and $S1, R1$ using SwinTransformer.

\begin{figure}[t]
\centering
\setlength\abovecaptionskip{2.5pt}   
\setlength\belowcaptionskip{2.5pt}   
\vspace*{-5pt}
\resizebox{0.95\textwidth}{!}{%
\includegraphics[width=\linewidth]{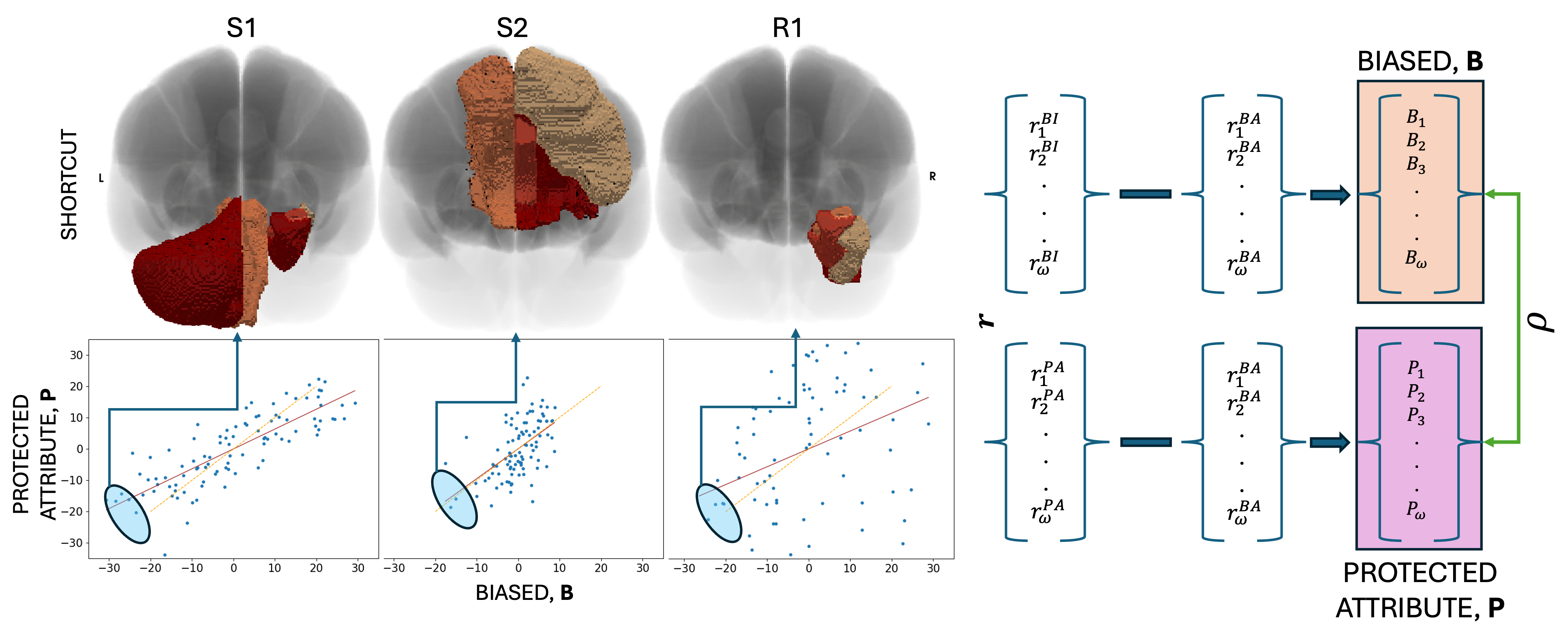}}
\caption{Experiment 3 - Interpretability. $S1$, $S2$ and $R1$ highlight the top-$5$ regions (shortcut features) with the highest biased $+$ protected attribute ranks ($\textbf{B}+\textbf{P}$). Negative indicates a higher rank and a darker shade of colour indicates more importance. The plots illustrate the linear relationships between biased and protected attribute ranks.}
\label{fig:explots}
\end{figure}

Next, we establish the stability of our explanations.
We selected the top $12.5\%$ patches with highest attributions and added brain patches with an increasing radius. We observed an average soft stability (computed over two different models) of about $0.95$ for all radii (with steps) up to $90$ (patches) for the ResNet50 model indicating high stability. For SwinTransformer, we observed an overall lower average stability of about $0.79$ and relatively more deviations. Therefore, in Figure~\ref{fig:explots}, we show the qualitative results for ResNet50 only for a more reliable interpretability analysis.

\section{Discussion}
\label{discussion}
Our work has shown that protected attributes (race and sex) can be classified to a high level of accuracy across multiple datasets and with two different DL models, one CNN based and one Vision Transformer based. 
This is an important prerequisite for the presence of bias in DL models and furthermore has the potential to lead to protected attribute based shortcuts. Given that many public brain MRI datasets do not report protected attribute information or are imbalanced, this raises concern for the use of DL for clinical classification purposes. 

Our baseline CN vs. AD classification model showed strong performance. However, this performance dropped significantly for the sex biased datasets. The test performance on majority classes (in training) generally increased and generally decreased from the baseline models on the minority classes. There was also a drop for the race biased model, although it was less pronounced. This is likely due to the less severe difference in imbalance in White and Black subjects between the training and test datasets. This was unavoidable as the total number of Black subjects (CN+AD) was relatively low. Additionally, the SwinTransformer model was trained from scratch, whereas ResNet50 used pre-trained weights, and this could have caused a steeper drop in Black AD classification due to a distributional shift. Essentially, due to the low number of samples from Black subjects with AD, the SwinTransformer model may not have been able to classify them correctly (see Figure~\ref{fig:diseaseplots}). However, a deeper analysis is required with more Black subjects to produce stronger evidence of shortcut learning and/or distributional shifts.

Qualitatively, it was initially difficult to find common patterns across models and tasks that could provide conclusive evidence that the biased models were utilising features related with the protected attributes. This is in contrast to commonly seen feature attributions for models trained on datasets like Waterbirds, where shortcut features like Background are clearly seen. 
Therefore, we performed rank based analysis to highlight the regions contributing to the shortcut learning.
This novel analysis enabled us to quantify the extent of the shortcut learning as well as reveal the regions that were associated with it, even though they are not ``visible'' to humans. This approach could be used to gain a deeper understanding of shortcut learning in other applications in medical imaging.

In summary, our work has shown that sex and race can be classified from 3D adult brain MRIs with high accuracy, representing the most comprehensive study of protected attribute identification from brain MRI to date. Sex classification (on adolescent brain MRIs) has previously been demonstrated in \cite{stanley2022fairness} but to the best of our knowledge, this is the first time that race classification from brain MRI has been investigated.
Furthermore, we have shown that protected attribute based (im)balance in the training set can cause shortcut learning and bias in AD classification. This is consistent with the findings of~\cite{wang2023bias}, who found that DL models for AD classification suffered from bias when trained with mixed (but imbalanced by sex and race) datasets. Additionally, another study~\cite{wang2024drop} showed that age and sex can be classified from Brain MRIs resulting in biased diagnosis and that augmentation strategies may aid in bias mitigation. All of these works examined bias when training with mixed (but imbalanced) training sets. In contrast, in our work we deliberately curated  biased datasets (similar to~\cite{sagawa2019distributionally} in the computer vision field) to create scenarios in which shortcut learning could be demonstrated and understood and show through multiple experiments that shortcut learning based on sex and race does occur. Finally, we also provide an approach to quantify the extent of shortcut learning and identify shortcut features using interpretability.


\begin{credits}
\subsubsection{\ackname} 
This research was supported by the UK Engineering and Physical Sciences Research Council (EPSRC) [Grant reference number EP/Y035216/1] Centre for Doctoral Training in Data-Driven Health (DRIVE-Health) at King's College London. 
Data used in preparation of this article were obtained from the Alzheimer’s
Disease Neuroimaging Initiative (ADNI) database (adni.loni.usc.edu). As such,
the investigators within the ADNI contributed to the design and implementation of ADNI and/or provided data but did not participate in analysis or
writing of this report. A complete listing of ADNI investigators can be found at:
\url{http://adni.loni.usc.edu/wp-content/uploads/how to apply/ADNI Acknowledgement List.pdf}.
Data were also provided in part by OASIS and the Human Connectome Project, 
WU-Minn Consortium (Principal Investigators: David Van Essen and Kamil Ugurbil; 1U54MH091657) funded by the 16 NIH Institutes and Centers that support the NIH Blueprint for Neuroscience Research; and by the McDonnell Center for Systems Neuroscience at Washington University.


\end{credits}
%
%
%
\bibliographystyle{splncs04}
\bibliography{references}
%
\end{document}